\documentclass[conference]{IEEEtran}
\IEEEoverridecommandlockouts
\usepackage{cite}
\usepackage{amsmath,amssymb,amsfonts}
\usepackage{algorithmic}
\usepackage{graphicx}
\usepackage{textcomp}
\usepackage{xcolor}
\usepackage{multirow}
\usepackage{subcaption}
\usepackage{cuted}
\usepackage{environ}
\NewEnviron{myequation}{%
\begin{equation}
\scalebox{0.9}{$\BODY$}
\end{equation}
}

\def\BibTeX{{\rm B\kern-.05em{\sc i\kern-.025em b}\kern-.08em
    T\kern-.1667em\lower.7ex\hbox{E}\kern-.125emX}}
\begin{document}

\title{Rep Smarter, Not Harder: AI Hypertrophy Coaching with Wearable Sensors and Edge Neural Networks\\}

\author{\IEEEauthorblockN{Grant King}
\IEEEauthorblockA{\textit{Computer Science and Engineering} \\
\textit{University of South Carolina}\\
Columbia, United States of America \\
gtking@email.sc.edu}
\and
\IEEEauthorblockN{Musa Azeem}
\IEEEauthorblockA{\textit{Computer Science and Engineering} \\
\textit{University of South Carolina}\\
Columbia, United States of America \\
mmazeem@email.sc.edu}
\and
\IEEEauthorblockN{Savannah Noblitt}
\IEEEauthorblockA{\textit{Computer Science and Engineering} \\
\textit{University of South Carolina}\\
Columbia, United States of America \\
snoblitt@email.sc.edu}
\and[\hfill\mbox{}\par\mbox{}\hfill]
\IEEEauthorblockN{Ramtin Zand}
\IEEEauthorblockA{\textit{Computer Science and Engineering} \\
\textit{University of South Carolina}\\
Columbia, United States of America \\
ramtin@cse.sc.edu}
\and
\IEEEauthorblockN{Homayoun Valafar}
\IEEEauthorblockA{\textit{Computer Science and Engineering} \\
\textit{University of South Carolina}\\
Columbia, United States of America \\
homayoun@cse.sc.edu}
% \and
% \IEEEauthorblockN{}
% \IEEEauthorblockA{\\
% \\
% \\
% }
}

\maketitle
\thispagestyle{plain}
\pagestyle{plain}

\begin{abstract}
    Optimizing resistance training for hypertrophy requires balancing proximity to muscular failure, often quantified by Repetitions in Reserve (RiR), with fatigue management. However, subjective RiR assessment is unreliable, leading to suboptimal training stimuli or excessive fatigue. This paper introduces a novel system for real-time feedback on near-failure states (RiR $\le$ 2) during resistance exercise using only a single wrist-mounted Inertial Measurement Unit (IMU). We propose a two-stage pipeline suitable for edge deployment: first, a ResNet-based model segments repetitions from the 6-axis IMU data in real-time. Second, features derived from this segmentation, alongside direct convolutional features and historical context captured by an LSTM, are used by a classification model to identify exercise windows corresponding to near-failure states. Using a newly collected dataset from 13 diverse participants performing preacher curls to failure (631 total reps), our segmentation model achieved an F1 score of 0.83, and the near-failure classifier achieved an F1 score of 0.82 under simulated real-time evaluation conditions (1.6 Hz inference rate). Deployment on a Raspberry Pi 5 yielded an average inference latency of 112 ms, and on an iPhone 16 yielded 23.5 ms, confirming the feasibility for edge computation. This work demonstrates a practical approach for objective, real-time training intensity feedback using minimal hardware, paving the way for accessible AI-driven hypertrophy coaching tools that help users manage intensity and fatigue effectively.
\end{abstract}


\section{Introduction}
Resistance training (RT) is a cornerstone of skeletal muscle hypertrophy, requiring precise balance between mechanical tension and fatigue management to maximize adaptive responses \cite{schoenfeld2010mechanisms}. A critical determinant of training efficacy is the proximity of sets to momentary muscular failure, commonly quantified through repetitions in reserve (RiR) \cite{refalo2024similar}. Studies find that terminating sets within 0--2 RiR optimizes hypertrophy while minimizing excessive fatigue \cite{refalo2024similar, izquierdo2006differential}. However, subjective self-assessment of RiR is often unreliable and leads to uncertainty, creating a fundamental challenge: undertraining by excessive RiR preservation limits hypertrophy stimuli, while overtraining through repeated failure impairs recovery.

Emerging wearable technologies offer unprecedented opportunities to objectify fatigue monitoring through inertial measurement units (IMUs) and edge computing. Prior work has demonstrated IMU-based estimation of perceived exertion \cite{albert2021using} and fatigue detection \cite{jiang2021data}, but real-time RiR prediction remains unexplored---a critical gap given its direct relationship to training decisions. Current approaches suffer from three key limitations: (1) dependence on supplementary sensors like ECG \cite{albert2021using} or force plates \cite{jiang2021data} that limit practicality, (2) post-hoc analysis rather than real-time feedback, and (3) absence of direct feedback on RiR for users to make informed decisions.

We propose a novel approach of real-time RiR feedback using a single wrist-mounted IMU, enabling users to optimize training sessions and avoid excessive fatigue. Our approach leverages a two-stage pipeline: (1) a segmentation model to detect the end of each repetition in real time, and (2) a classification model to predict near-failure repetitions (RiR $\leq$ 2). This work aims to advance the field of wearable sensor-based training feedback by providing real-time RiR predictions, enhancing training efficiency, and minimizing fatigue.

\subsection{Background}
In the context of resistance training, momentary muscular failure is the point during a set at which the lifter can no longer complete a repetition with proper form. Reps in Reserve (RiR), then, is the measure of how many repetitions a lifter could have completed beyond the current set. For example, if a lifter completes 8 reps with a weight they could have lifted for 10 reps, they are said to have 2 RiR. This measure is often used to gauge the intensity of a set and can help lifters determine when to stop a set to avoid excessive fatigue. 

Taking sets to near-momentary muscular failure is often the target of RT, as it provides the most effective stimulus for muscle growth. However, training to complete failure can lead to excessive fatigue and hinder recovery without much benefit. It is therefore important to balance the benefits of training to failure with the need for adequate recovery. Adjusting RT based on RiR can be beneficial across various training populations. Bodybuilders training for hypertrophy, for example, may choose to leave a few repetitions in reserve to reduce accumulated fatigue while still maintaining progress. Likewise, athletes training for strength and power can strategically manage RiR to optimize performance outcomes \cite{refalo2024similar, izquierdo2006differential}. Lastly, incorporating RT into physical therapy has demonstrated positive effects on muscle function and recovery in rehabilitation settings \cite{kristensen2012resistance}. In such cases, adjusting repetitions in reserve plays a key role in managing the balance between promoting progress and avoiding reinjury.

\subsection{Related Work}
Previous works in resistance training monitoring have made significant advancements. Although real-time prediction of RiR remains an under-explored area, recent studies have explored estimating perceived exertion (RPE), detecting fatigue during resistance training, and classifying lifting quality using wearable sensors. 

Prior studies have investigated using wearable sensors for resistance training monitoring. For instance, Albert et al. \cite{albert2021using} explored predicting RPE (considered to be the inverse of RiR) using machine learning with physiological and motion data from six IMUs and one ECG sensor during squats. They designed several models to make post-hoc RPE predictions of resistance training sets, achieving a best Pearson correlation of 0.89 with a Gradient Boosted Regression Trees (GBRT) model. Similarly, Jiang et al. \cite{jiang2021data} modeled per-repetition fatigue onset across various exercises using multimodal sensor data, including 17 IMUs and a force plate. In this study, a CNN model outperformed a Random Forest model, scoring Pearson correlation coefficients of 0.89-0.94 across the exercises. Elshafei and Shihab \cite{elshafei2021towards} conducted a study closely related to our own, focused on detecting bicep muscle fatigue during concentration curls using a wrist-mounted IMU and an Apple Watch for heart rate monitoring, achieving high accuracy with a two-layer Feed-forward Neural Network. A key distinction here is the prediction of fatigue---a decrease in the mechanical capabilities of a muscle---rather than muscular failure, as we explore. These studies provide key insights into the feasibility of sensor-based fatigue and exertion estimation using wearable sensors and various machine learning techniques and models. However, their reliance on supplementary sensors (such as the force plate and large number of IMUs) and/or post-hoc analysis limits their applicability for portable, real-time RiR prediction. 

Other research has touched upon qualitative activity recognition and muscle fatigue classification using different sensor modalities. Velloso et al. \cite{velloso2013qualitative} assessed weightlifting exercise quality to detect mistakes in exercise form using an array of IMU sensors, achieving high accuracy with Random Forest models. Dang et al. \cite{dang2023fatigue} investigated muscle fatigue classification during force-relaxation cycles using surface electromyography (sEMG) signals, with an Attention-LSTM model showing strong performance. Although not directly applicable to prediction of muscular failure, the techniques and models used in these studies provide valuable insights into the feasibility of sensor-based fatigue and exertion estimation.

While prior research has addressed fatigue classification, perceived exertion estimation, and lifting quality, to our knowledge, no existing study has proposed a complete, wearable sensor-driven pipeline for real-time prediction of repetitions in reserve. This remains an open area of research with substantial opportunity for further exploration and advancement.

\subsection{Contributions}
In this work we explore the novel task of developing models of human activity in the gym and predicting RiR in real time using wearable, easy-to-apply sensors. As a first step toward this goal, we propose the development of a neural network-based pipeline for segmenting bicep curl reps and detecting ``near-failure'' repetitions in real time using a wrist-mounted IMU. Near-failure reps refer to repetitions performed with two or fewer remaining repetitions in reserve (RiR $\leq$ 2). Such a model will allow for real-time feedback to be provided to users, helping them to optimize their training sessions and stop their sets with at most two RiR. The contributions of this work are as follows:

\begin{itemize}
    \item We introduce a new dataset of wrist IMU data collected during preacher curls performed to momentary muscular failure by 13 diverse participants. The dataset includes over 600 annotated repetitions with varied form, weight, and number of reps.
    \item We develop a segmentation model to detect the end of each repetition in real time.
    \item We develop a classification model to identify near-failure repetitions.
    \item We demonstrate that our models can be deployed on edge hardware using a Raspberry Pi 5 and Apple Watch Series 10, validating their potential for real-time, low-power operation in various gym settings.
\end{itemize}

\section{Materials and Methods}

\subsection{Data Collection}

\subsubsection{Participants}
Participants for data collection consisted of this work's authors and volunteers at three different gyms. There were 13 participants in total (9 male, 4 female). All participants were about 18-25 years old and varied in weight, height, strength, and resistance training experience.

\subsubsection{Equipment}
\begin{figure}[]
    \centering
    \includegraphics[width=0.5\linewidth]{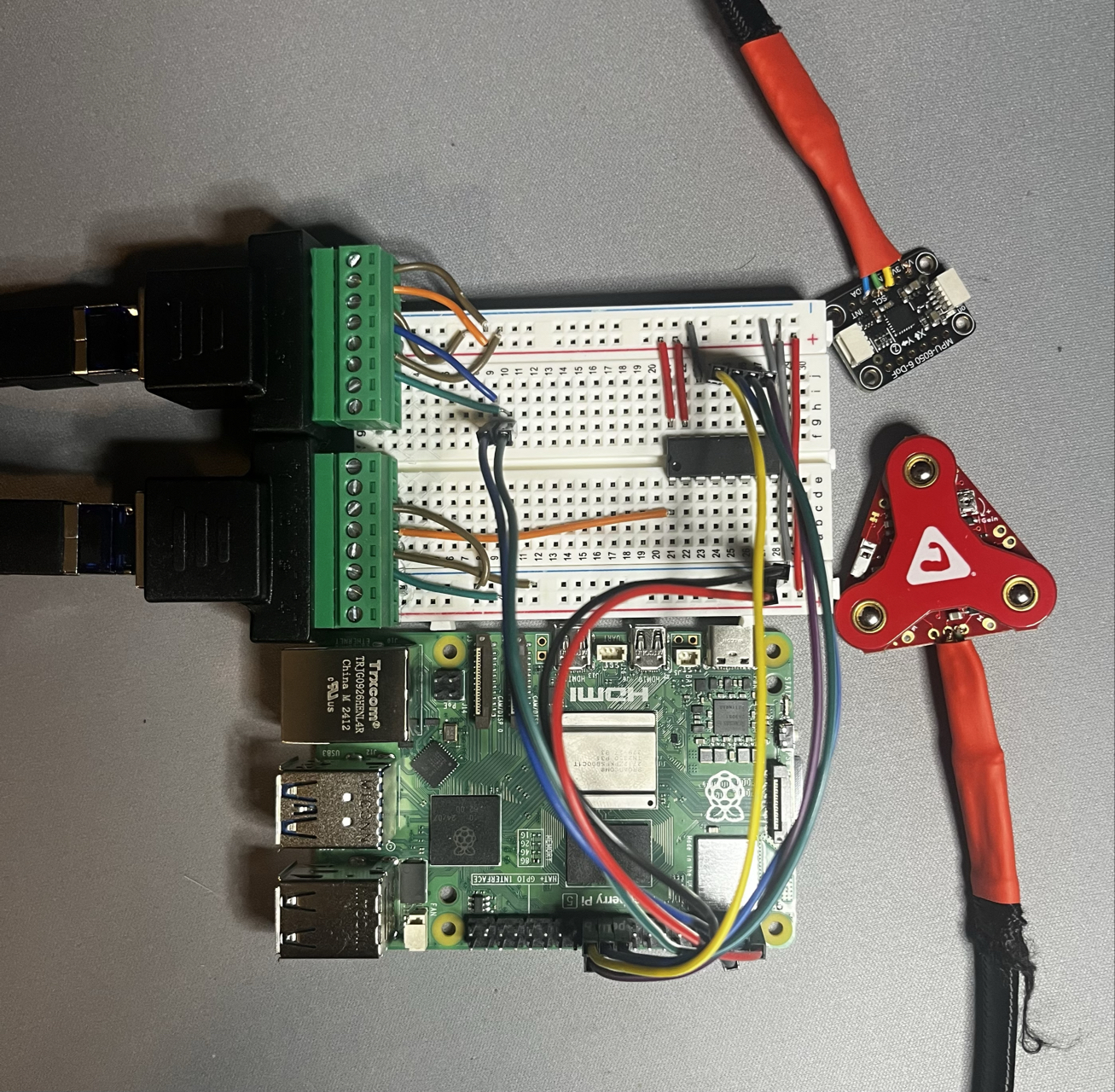}
    \caption{\textbf{Data collection device.} The Raspberry Pi (bottom) reads, processes, and stores data from the IMU (top right). %The sEMG sensor (bottom right) was not used due to hardware issues. 
    During data collection, the Pi was powered by a portable battery pack. A box was used to protect the Raspberry Pi from damage, and the IMU was taped to participants' wrists.}
    \label{fig:data_collection_tool}
\end{figure}

Participant preacher curls were recorded using a Raspberry Pi 5 and the Adafruit MPU-6050 6-axis IMU, capturing both accelerometer and gyroscope data. Although only IMU data was used here, the device, shown in Figure \ref{fig:data_collection_tool}, was designed with the intention and capability of recording data from multiple sensors (such as heartbeat sensor and sEMG). The Raspberry Pi reads data from the IMU, and was connected to and accessed by a laptop. Three different preacher curl machines at the three locations were used for data collection.

\subsubsection{Procedure}
\begin{figure}[]
    \centering
    \includegraphics[width=0.5\linewidth]{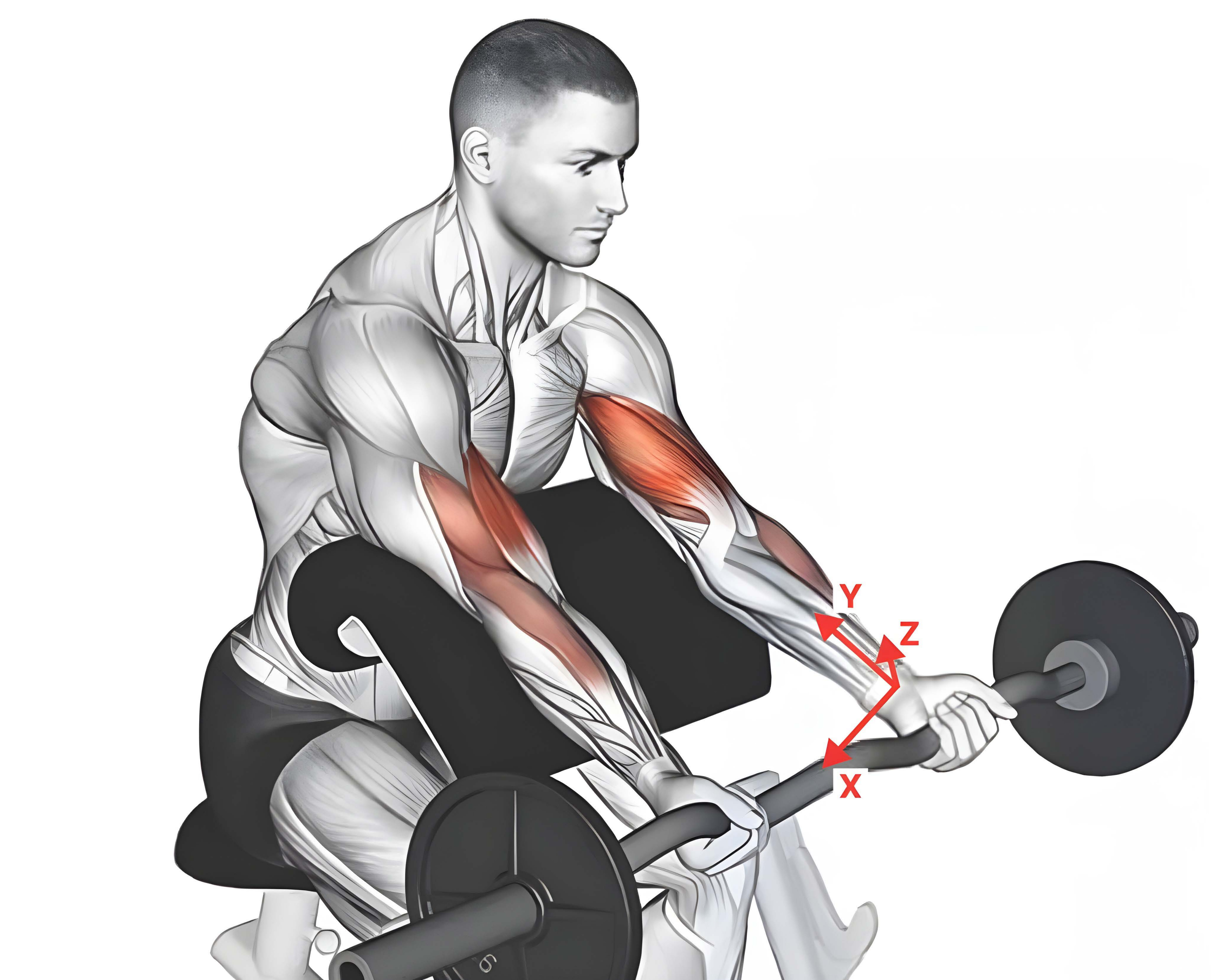}
    \caption{\textbf{IMU orientation.} The IMU was taped to the participant's wrist with the axes oriented as shown. Image source \cite{preacher_curl}.}
    \label{fig:imu_orientation}
\end{figure}

During data collection, the IMU was taped to each participant's left wrist. The axes of the sensor were oriented as shown in Figure \ref{fig:imu_orientation}. Each participant was instructed to perform sets of preacher curls with consistent form, taking each set to momentary muscular failure. Participants were free to select the weight they lifted and decide how many sets they performed. Consequently, collected data varied in the amount of weight lifted, the number of repetitions per set, and how many sets were completed in a session. Participants also introduced natural variation in form and range of motion (e.g. locking out their arms at the bottom of the curl, duration of pause at the bottom, height of the concentric phase). Only sets taken to failure with proper form were admitted to this work's dataset.

During data collection, annotations were recorded for future labelling. This was facilitated by a custom Python server running on the Raspberry Pi. When prompted to start recording, the server reads and stores data from the IMU in CSV format at approximately 100 Hz. The server broadcasts an interactive data collection web application for access via laptop over Wifi or Ethernet. This data collection dashboard was used to enter participant and location metadata, start the recording of IMU data, manually record timestamp markers with a button to annotate the end of each repetition, and stop the recording. The timestamp markers were later used to generate data labels.

In total, 13 participants performed 68 sets of preacher curls, with an average length of 36 seconds and 9.3 curl repetitions. This resulted in 40 minutes of data consisting of 631 reps.

\subsection{Data Processing}
Several steps were taken to process the raw IMU data and prepare it for model training. 

\begin{figure*}[h!]
    \centering
    \includegraphics[width=0.85\textwidth]{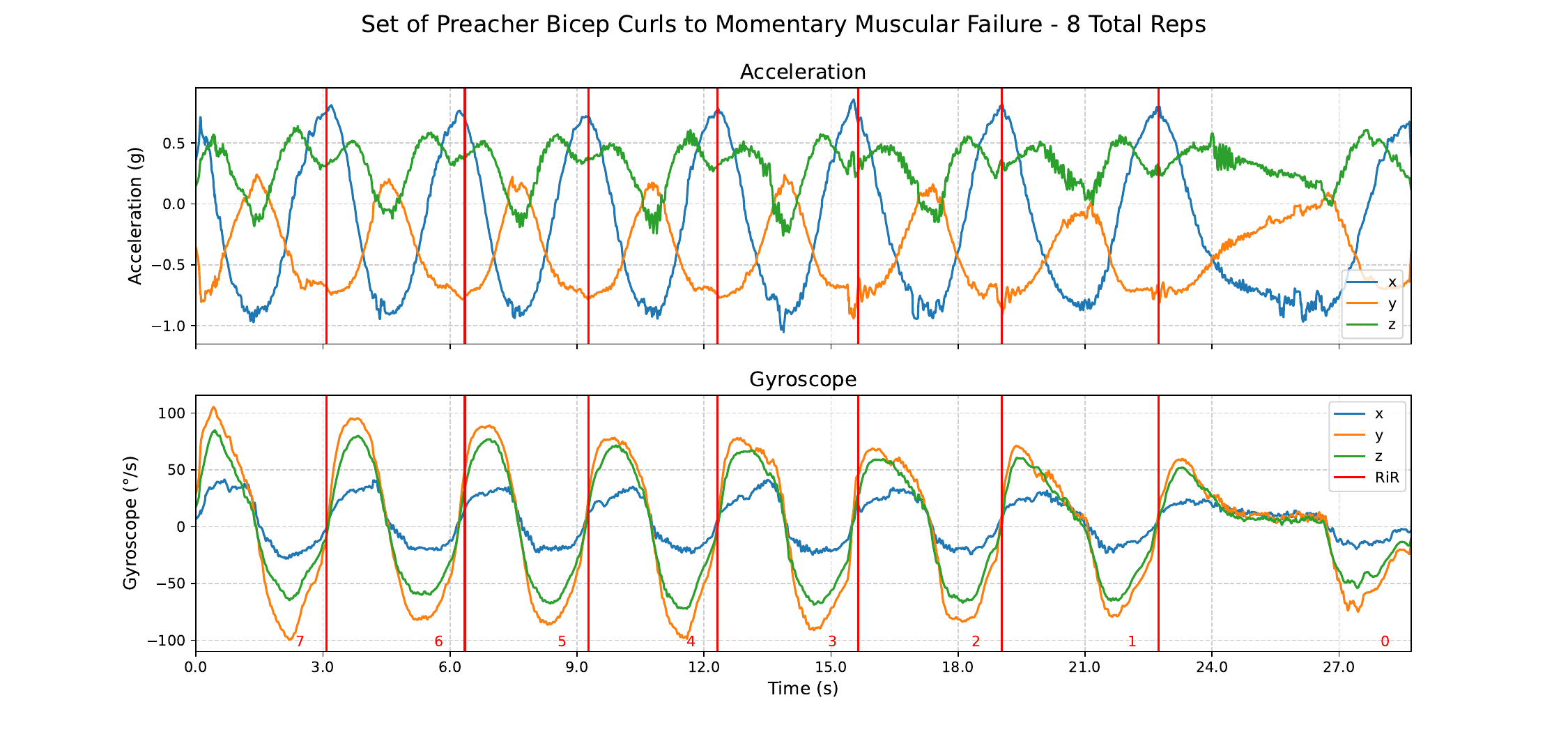}
    \caption{\textbf{Full set of preacher curls.} The accelerometer (top) and gyroscope (bottom) data from a single set of preacher curls, with the end-of-rep annotations shown as red vertical lines. The set lasted 28.7 seconds and consisted of 8 repetitions to failure. Red numbers indicate the Repetitions in Reserve at each point during the set. The last rep, at which point the participant could no longer lift the weight, is a rep taken to failure, with an RiR of 0.}
    \label{fig:full_set}
\end{figure*}

\subsubsection{Data Cleaning}
First, to resolve fluctuations in data sampling rate caused by the Raspberry Pi's CPU load, the 6-channel timeseries data was first interpolated to exactly 100 Hz. Next, the data was smoothed with a 15-data point (150 ms) moving average filter to remove signal noise. The data annotations were adjusted as well. Manual inspection of the timeseries signals was used to adjust any incorrect end-of-rep annotations to the true end of the rep. A full set of curls after cleaning, along with the data annotations, can be seen in Figure \ref{fig:full_set}.

\subsubsection{Splitting \& Windowing}
To facilitate model training, the collected data was split into train and validation splits. About 80\% of the recorded preacher curl sets (53) were added to the train set. The remaining 20\% of sets (15) were added to the validation split. Next, the data in each split was windowed into 256 data-point (2.56s) segments, with a stride of 2 data points (20 ms). Each window was labelled based on the end-rep-marker annotations, as will be explored. In total, about 90k training windows and 26k validation windows were generated.

Splitting data at the set level guarantees that no overlap exists between the data in the training and validation set. %, ensuring that the model is not trained on data it will be evaluated on. 
Windows of 2.56 seconds were chosen to balance the receptive field of the model with real time capabilities. Windows of this length capture entire reps, which are usually less than 2.56 seconds, while maintaining support for making real time predictions at up to about 0.4 Hz. A small stride was chosen to maximize the number of training samples. %available to the model. 

\subsubsection{Data Augmentation}
%To simulate additional data, 
The training data was augmented with several techniques. First, training samples were randomly stretched and cropped in the time dimension by a factor of 1 to 1.5, simulating faster reps. Next, the amplitude of the training windows were randomly scaled by a factor of 0.6 to 1.4, simulating more or less forceful reps. Augmentations were randomly applied to the training windows on every retrieval.

\subsection{Segmentation Model Design}
\begin{figure}[t]
    \centering
    \includegraphics[width=0.9\linewidth]{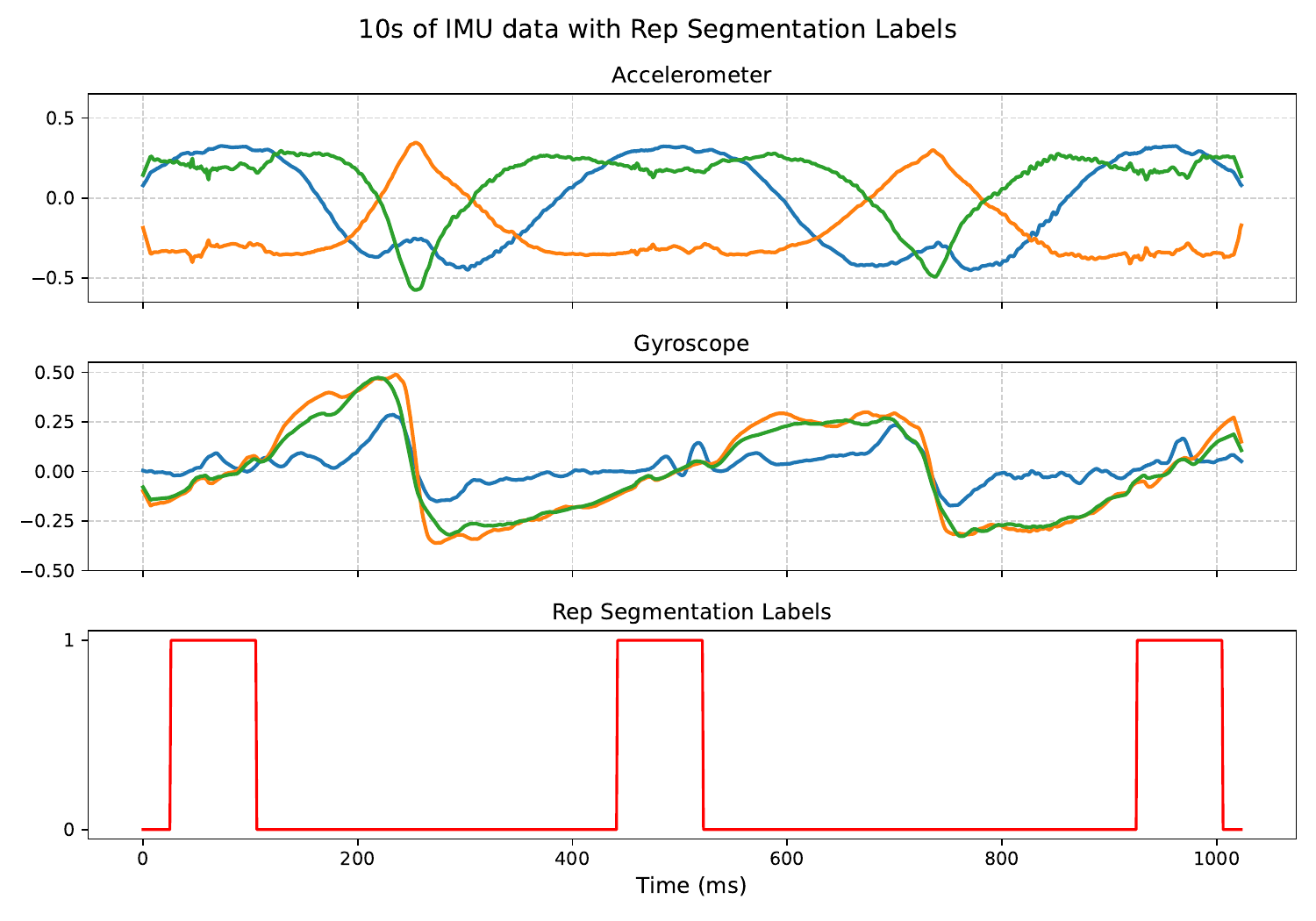}
    \caption{\textbf{Segmentation labels.} A single window of 1024 data points. A larger window size is chosen here for demonstration purposes. Shown from top to bottom are the acceleration data, gyroscope data, and segmentation labels. The segmentation labels are binary, with 1 indicating the end of a rep.}
    \label{fig:seg_labels}
\end{figure}

The first stage of the pipeline proposed in this work is a segmentation model, designed to detect the end of each rep in a set of bicep curls. This model was designed to take as input a window of 6-channel IMU data and output a binary classification of whether each input data point is the end of a rep. To train the model, each window was given 256 labels (one per data point) based on the end-rep-markers. The area (80ms) around any end-rep-markers occurring within a window was labelled with a positive label. The remaining data points of each window were given negative labels. An example of the labelling for a single window is shown in Figure \ref{fig:seg_labels}. 

\subsection{Near-Failure Classification Model Design}
The next step in this work was to design a classification model to predict near-failure reps. This model was designed to classify a window of IMU data as either a near-failure rep (RiR $\le$ 2) or not. The model was trained on the same data as the segmentation model, but with a different set of labels. Each window was given a binary label indicating whether the entire window was a near-failure rep or not. Windows with over 50\% of their data points in a region of RiR $\le$ 2 (for instance, the region corresponding to the last three reps in Figure \ref{fig:full_set}) were given a positive label. The remaining windows were given a negative label.

\subsection{Model Architectures}
\subsubsection{Segmentation Model}
\begin{figure}[]
    \centering
    \includegraphics[width=0.9\linewidth]{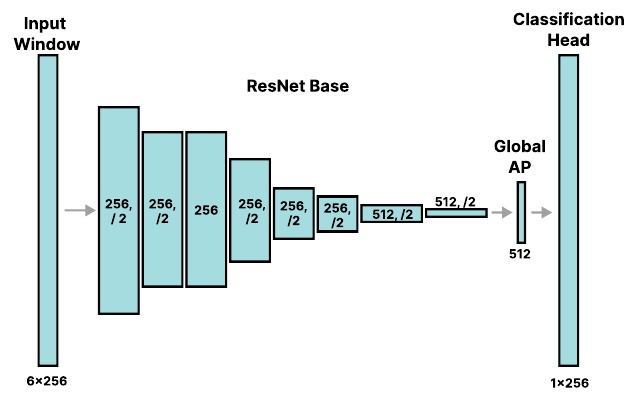}
    \caption{\textbf{Segmentation model architecture.} Each block in the ResNet Base consists of a 1D Convolution layer with a kernel size of 3, followed by a batch normalization layer and a ReLU activation function. Residual connections are included around each block. Labels on each block represent the output channels and stride of each layer.}
    \label{fig:seg_arch}
\end{figure}

The foundation of the segmentation model is a 1D convolutional residual neural network (ResNet) \cite{he_2016}. The model consists of several strided 1D convolutional layers, each followed by a batch normalization layer and a ReLU activation function. Residual connections are included around each convolutional layer, with projection layers included as necessary, as in \cite{he_2016}. A final global average pooling and linear classification layer produce 256 output confidences, one for each data point in the input window. The final model architecture---found by neural architecture search (NAS) \cite{zoph_2016} as explored in the next section---consists of 2,971,648 parameters and is shown in Figure \ref{fig:seg_arch}.

\begin{figure}[t]
    \centering
    \includegraphics[width=0.9\linewidth]{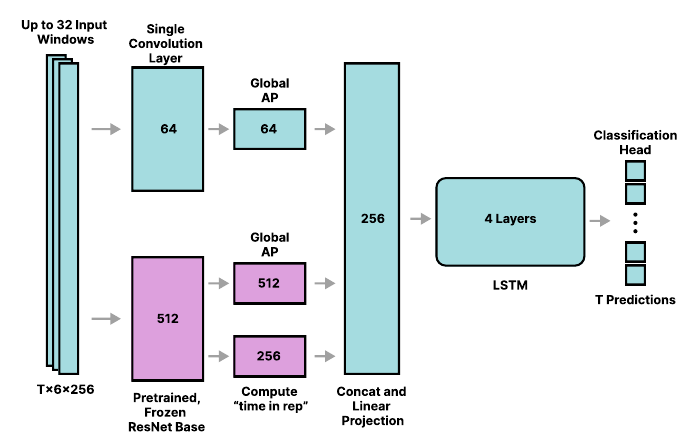}
    \caption{\textbf{Classification model architecture.} }
    \label{fig:class_arch}
\end{figure}

\subsubsection{Near-Failure Classification Model}
The near-failure classification model integrates the trained ResNet base and classification head of the segmentation model in its architecture. The weights of the trained segmentation model are loaded and frozen (not updated during training). The previous model is then integrated as follows:
\begin{itemize}
    \item ResNet Base: serves as a feature extractor, providing high-dimensional representations of the input data. The output encodings of the ResNet base after global average pooling (vector of 512 values) are used in the next stage of the classification model.
    \item Segmentation Outputs: The output predictions of the segmentation model are also used in the classification model. The segmentation model predicts any end-of-rep occurrences within a window. After smoothing and finding the midpoints of positive prediction regions, the resulting end-of-rep marker predictions are used to calculate the time spent in each rep that occurs in a window. A vector of 256 values is generated, with each value representing the time spent in the rep occurring at that data point. This vector is referred to as ``time-in-rep'', and is also used in the next stage of the classification model.
\end{itemize}

The trainable portion of this model consists of a single convolution layer, a linear projection layer, a one-directional LSTM component, and a classification head. This model includes a temporal component, and accepts $T$=1 to 32 concurrent windows of data (with a stride of 64) as input, making a prediction for each timestep. All layers besides the LSTM component operate on these multiple input windows in parallel with no communication between them (as if there were a second batch dimension). The hyperparameters of these components were also found through NAS. The model is structured as follows:
\begin{itemize} 
    \item Skip Convolution Layer: provides a direct path from the input signals to the projection layer, allowing the model to optionally bypass the ResNet Base. In the final model architecture, this layer has a kernel size of 3, 64 output channels, and a stride of 1.
    \item Linear Projection Layer: consolidates the outputs of the ResNet Base (512), the time-in-rep vector (256), and the outputs of the skip convolution layer after global average pooling (64). This layer has $512+256+64=832$ input channels and output channels equal to the hidden size of the LSTM layers.
    \item LSTM Component: operates over multiple windows of data, allowing the model to learn temporal dependencies. The LSTM is one-directional, meaning it only processes the input data in one direction (past to future). Given an input of $T$ windows, the LSTM component outputs $T$ hidden states, representing the encodings of each input window to be used for classification. In the final architecture, the LSTM component has a hidden size of 256 and consists of 4 layers.
    \item Classification Head: linear layer with 256 input features and 1 output. This layer takes the final hidden states of the LSTM component for each window and produces a single prediction confidence for each.
\end{itemize}

The final model architecture is illustrated in Figure \ref{fig:class_arch}. It consists of 5,291,841 parameters, of which 2,320,193 are trainable.

\subsection{Neural Architecture Search}
Neural architecture search (NAS) was used to find the optimal hyperparameters of the segmentation and classification models. NAS was implemented using the Optuna library \cite{akiba_2019}. Each optimization ran for 100 trials and was configured to maximize the F1 score of the model on the validation set. The initial search for the segmentation model architecture included the optimization of the number of ResNet Block Stages (a set of convolutional layers with the same number of output channels and a stride of 2 in the first layer) [4 to 8], the number of ResNet Blocks per stage [1 or 2], the number of output channels of the first and last stages [64 to 256 and 128 to 512], and the kernel size of the first convolution layer [3,5, or 7] within the specified ranges. During these trials, the number of output channels of all intermediate stages were log-linearly interpolated to increase from the output channels of the first stage to that of the final stage. Based on the initial search results for the segmentation model, a second search was performed to further optimize the learning rate and weight decay of the model optimizer.

Next, the hyperparameters were searched once for the classification model. The hyperparameters optimized for the classification model included the number of output channels of the skip convolution layer [32 to 512], the number of LSTM layers [2 to 5], and the hidden size of the LSTM [32 to 512].

During optimization, all models were trained until the validation F1 score showed no improvement for 15 epochs. The models resulting from trials with the highest validation F1 scores were selected as the best models for each task. In the case of trials with similar performances, models with lower parameter counts were selected.

\subsection{Model Training}
Models were trained using the AdamW optimizer with learning rates of 4.3e-3 (segmentation model) and 7.9e-4 (classification model). The segmentation model was optimized with a combination of Binary Cross Entropy and Mean Squared Error loss, as shown in Equation \ref{eq:seg_loss}, where $\alpha=0.8$. The classification model was optimized with Binary Cross Entropy loss. The models were trained for 50 epochs or until validation F1 score stopped improving (resulting in 35 epochs for the segmentation model and 4 for the classification model). Both models were trained with a batch size of 128 on a single NVIDIA RTX 4090 GPU using the Pytorch library \cite{paszke_2019}. 

\begin{myequation}%
Loss = \frac{1}{N} \sum_{i=1}^{N} \left( \alpha\cdot\text{BCE}(y_i, \hat y_i) + (1-\alpha)\frac{1}{256} \sum_{j=1}^{256} (\hat y_{ij} - y_{ij})^2 \right) %
\label{eq:seg_loss}
\end{myequation}

The segmentation model was trained on batches of individual windows of data from the train dataset. The classification model, on the other hand, was trained on batches of multiple windows of data. Each element of a batch was a set of $T=32$ concurrent windows, with a stride of 64. The total length of each of these sequences, accounting for overlap, is 2240 data points (22.4 seconds). For each sequence, the model was trained to predict the class of each window in the sequence using an LSTM with information flowing only from past windows to the future. In this way, the model was trained to make predictions using 1 up to 32 windows of data. It was also validated in this way, ensuring the model's performance was not reliant on having all 32 windows available at once. This is important for real-time inference, as the model must be able to make predictions on each window as it arrives.

\subsection{Model Evaluation}\label{sec:model_evaluation}
During training, models were validated using the F1 score and loss on the validation dataset windowed with a stride of 2. This ensured results between train and validation datasets were comparable during development. For the final evaluation, the same sessions that make up the validation set were used to evaluate the models. However, for this portion, each session was windowed with a stride of 64 data points (640 ms), simulating real-time inference. %After waiting 2.56 seconds to collect the first window of data, one prediction is made every 640 ms. 

In the case of the segmentation model, the model made predictions on each window independently, and the overlapping predictions were averaged to produce a single prediction for each data point in a session. The final predictions over each whole session were evaluated against the true segmentation labels to calculate the per-session F1 score. The final F1 score was calculated as the average of the per-session F1 scores.

\begin{figure}
    \centering
    \includegraphics[width=0.6\linewidth]{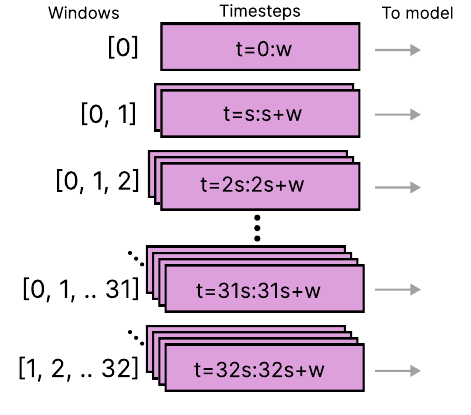}
    \caption{\textbf{Real-time inference.} Windows of size $w = 256$ are collected with a stride of $s = 64$ data points. The most recent $T=32$ (or less if not available) windows of data are used to make a prediction. The model makes $T$ predictions, and the prediction for the most recent window is used for real-time inference.}
    \label{fig:real_time_inference}
\end{figure}

For the classification model, the model made predictions using up to 32 windows. As illustrated in Figure \ref{fig:real_time_inference}, at the beginning of each session the model waits for 2.56 seconds for the first window of data to be collected, simulating real-time inference. After that, the model makes a prediction every 640 ms, using the most recent 32 windows of data (or less if not enough windows are available). The model's predictions for the most recent window at each inference are saved. %The final predictions of each session are evaluated against the true labels of each window to calculate the per-session F1 score, as before. The final F1 score is calculated as the average of the per-session F1 scores.

\subsection{Edge Deployment}
\subsubsection{Raspberry Pi}
To deploy the model on the edge, the preexisting data collection server on the Raspberry Pi was adapted to perform inference on a live set. After starting a session, 256 IMU samples are saved to build the first input window for the model. After that, each new window is created by discarding the most recent window's oldest 64 samples and appending the most recently recorded 64 samples from the IMU. These input windows accumulate to a max of 32, after which they are refreshed by new windows in a similar manner.

\subsubsection{iPhone 16 and Apple Watch Series 10}

% \begin{figure}[b]
%     \centering
%     \includegraphics[width=.4\linewidth]{images/watch}
%     \caption{\textbf{Orientation of Apple Watch axes.}
%     Image source \cite{watch}.}
%     \label{fig:watch}
% \end{figure}

Deploying to a real mobile device and companion wearable device required conversion of the model to Apple's Core ML representation, the creation of connected phone and watch applications, and the reconciliation of hardware differences. The orientation of the Apple Watch's axes %, pictured in Figure \ref{fig:watch}, 
was reconciled with the orientation of the IMU used during data collection by using the watch's positive $y$-axis, negative $x$-axis, and negative $z$-axis as positive $x$, $y$, and $z$-axes, respectively (accounting for the rotation of the user's wrist when using the curl machine).

The watch application collects and windows data as done on the Raspberry Pi. Due to the transfer size constraint, the watch application sends eight windows at a time to the phone. The phone application receives these windows to build the inference input of 32 windows, then maintains this buffer with new windows. The phone sends its predictions back to the watch for the user to receive haptic feedback upon positive predictions (ie. they are near failure).

\section{Results}
\begin{figure}
    \centering
    \begin{subfigure}[b]{0.3\textwidth}
        \includegraphics[width=\textwidth]{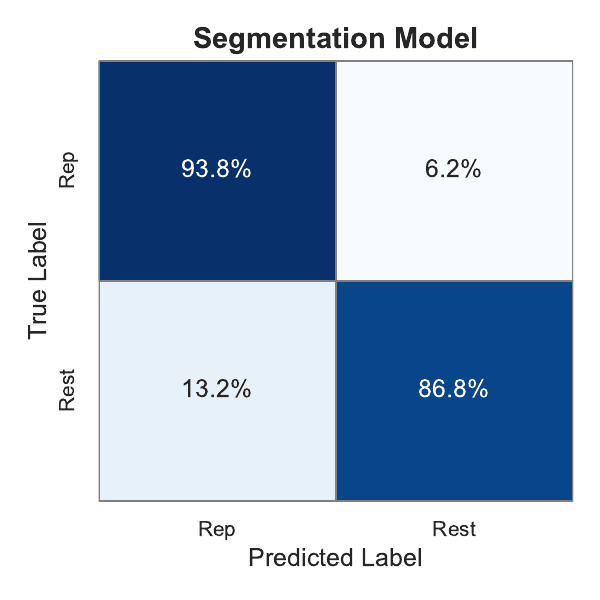}
    \end{subfigure}
    \hfill
    \begin{subfigure}[b]{0.3\textwidth}
        \includegraphics[width=\textwidth]{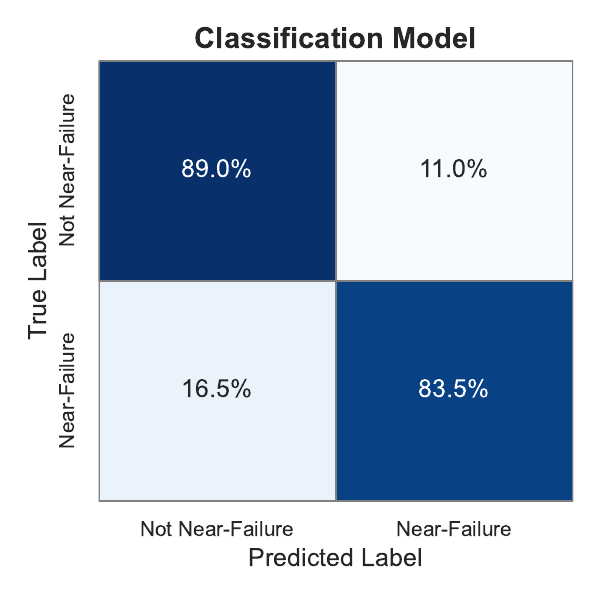}
    \end{subfigure}
    \caption{Confusion Matrices for the segmentation (above) and classification (below) models on the validation sessions. Values are normalized over the true labels (rows). For segmentation, "Rep" represents the time spent during a rep, and "Rest" is the time between reps. The labels for the classification matrix describe whether or not a rep is near-failure (RiR $\le$ 2).}
    \label{fig:cms}
\end{figure}

The segmentation and classification models were evaluated on the validation sessions, simulating real-time inference as described in Section \ref{sec:model_evaluation}. The segmentation model achieved an average F1 score of 0.83 and accuracy of 92.8\%. The classification model achieved an average F1 score of 0.82 and accuracy of 86.6\%. Table \ref{tab:eval} summarizes the average performances of the models over all the sessions, and the confusion matrices are shown in Figure \ref{fig:cms}.

\begin{table}[]
    \centering
    
    \renewcommand{\arraystretch}{1.5}
    \begin{tabular}{c|c|c|}
        \cline{2-3}
        & Segmentation Model  & Classification Model  \\ \hline
        \multicolumn{1}{|c|}{F1 Score}  & 0.830        & 0.824          \\ \hline
        \multicolumn{1}{|c|}{Precision} & 0.800        & 0.860          \\ \hline
        \multicolumn{1}{|c|}{Recall}    & 0.869        & 0.856          \\ \hline
        \multicolumn{1}{|c|}{Accuracy}  & 92.7\%       & 86.6\%         \\ \hline
    \end{tabular}
    \caption{\textbf{Model performance.} The average F1 score, precision, recall, and accuracy of the segmentation and classification models over each session in the validation dataset.}
    \label{tab:eval}
\end{table}

\subsection{Edge Deployment Performance}
\begin{figure}[]
    \centering
    \includegraphics[width=1\linewidth]{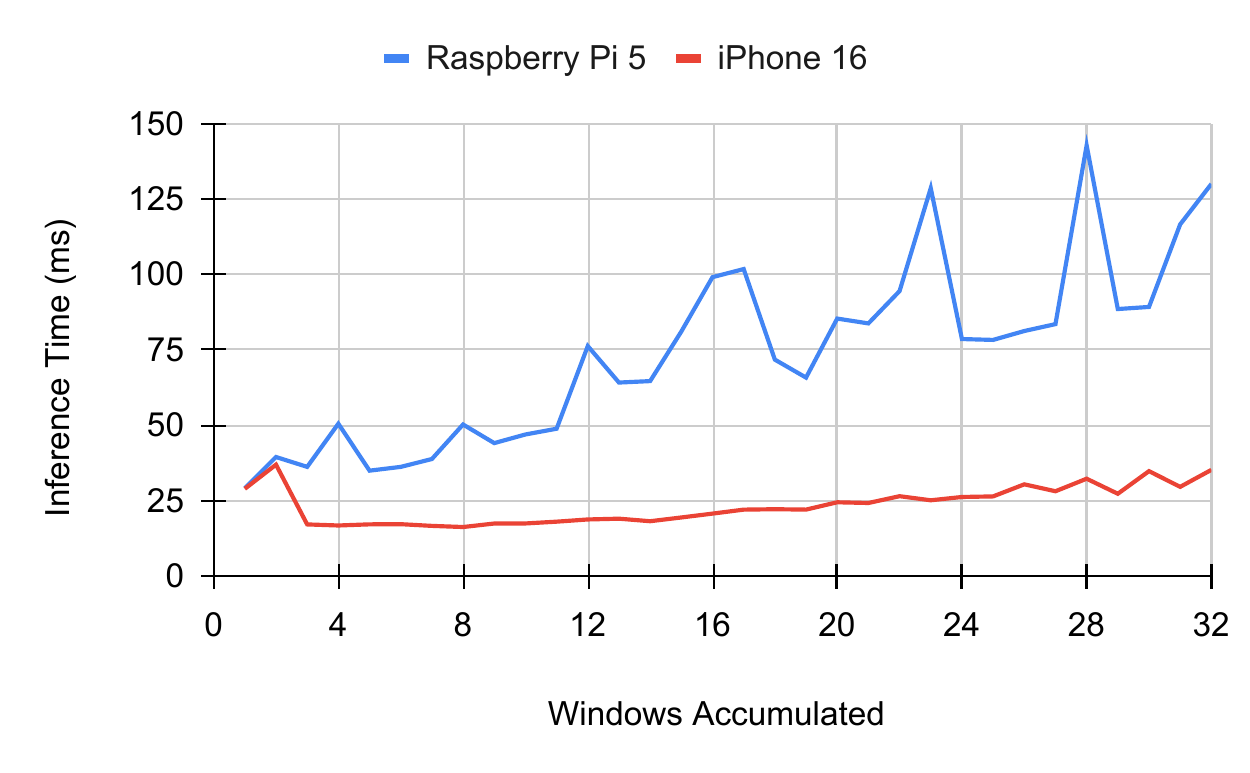}
    \caption{\textbf{Deployed model latency on Raspberry Pi and iPhone.} Inference latency ranging from one-window input to 32 windows.}
    \label{fig:deployed_model_latency}
\end{figure}

The final model was 20Mb in size, and was deployed on the Raspberry Pi for inference. The model runs inference on the growing list of windows, from length 1 to length 32. The average inference latency therefore increases as more windows are accumulated and finally settles once the max of 32 is reached. On the Raspberry Pi 5, the model performed inference throughout a live set with an average latency of 112.01 milliseconds (130.03 for 32-window inputs), visualized in Figure \ref{fig:deployed_model_latency}. Using PyTorch profiling, the convolution operator and the RNN operator cost 45.81 percent and 26.63 percent of inference CPU time, respectively.

The iPhone 16 deployment saw an average inference latency of 23.5 milliseconds (35.19 for 32-window inputs). Of that, 14.05 milliseconds were spent in the encoder, 1.51 milliseconds in the now-separated "time-in-rep" function, and 7.94 milliseconds in the LSTM, on average.

\section{Discussion}

\subsection{Key Achievements and Implications}
This study demonstrates that a single wrist-mounted IMU can effectively support real-time detection of resistance training repetitions and near-failure states. These models achieved high accuracy using only 6-axis inertial data, eliminating the need for complementary sensors like ECG or force plates used in prior work \cite{albert2021using, jiang2021data}. This reduces hardware complexity and cost while maintaining accuracy---a critical advantage for practical adoption in gym environments. 

The segmentation model achieved robust rep segmentation, with an F1 score 0.83 and recall of 0.87. The high recall of this model suggests that it is unlikely to miss the end of a rep, which is critical for accurate rep counting. The classification model also achieved high performance, successfully identifying near-failure repetitions (RiR $\le$ 2) with an F1 score of 0.82 and precision and recall of 0.86. This model is capable of providing actionable feedback to users without relying on subjective self-assessment. With a high precision, the model is unlikely to misclassify a rep as near-failure when it is not, which is important for ensuring that users do not stop their sets too early.

The segmentation model's real-time rep counting capability enables precise tracking of training volume, a cornerstone of hypertrophy programming. Integration of this model with applications, in a smart watch, for instance, could count reps for user convenience, circumventing the need for them to manually record their training. Our near-failure classifier offers a unique solution to the challenge of training intensity management. By alerting users when they are approaching failure, users are able to approach sufficient mechanical tension to stimulate growth, while preventing excessive fatigue that could impair recovery \cite{izquierdo2006differential}. 

Deployment on a Raspberry Pi 5 (inference latency: 112 ms, model size: 20 MB) validates the system's suitability for low-power, portable applications. Unlike cloud-dependent solutions, edge processing ensures privacy-preserving operation in environments with limited connectivity, such as commercial gyms or home setups. The 640 ms update stride (1.56 Hz) and low model latency of 112 ms provides timely feedback without overwhelming users, striking a balance between responsiveness and computational efficiency.

Deployment on iPhone 16 and Apple Watch Series 10 ensured timely predictions (23.5 ms average inference latency) on realistic target devices. The approach of running inference on the more capable phone results in this low latency as well as lower battery usage on the watch, but introduces application complexity.

\subsection{Limitations and Future Directions}
While promising, the study has limitations. First, data collection focused solely on preacher curls. Future work should validate generalizability to compound movements (e.g., squats, bench presses) where fatigue manifests differently \cite{velloso2013qualitative}. Second, our models were trained on a limited number of participants (13), which may not capture the full spectrum of lifting styles and fatigue responses. Expanding the dataset to include diverse populations, such as older adults or individuals with disabilities, is essential for broader applicability. Third, instead of using the Raspberry Pi, collecting training data with an Apple Watch directly or other target wearable would reduce data collection complexity and improve deployment accuracy.

Future work will involve a more diverse dataset, including additional exercises and participants, to enhance model generalizability. We also plan to explore the integration of additional sensors (e.g., stretchable strain sensors) to improve performance and robustness. The goals of the models will also be advanced in future work. Ideally we aim to develop a model that can predict RiR directly in real time, providing users with immediate feedback on their training intensity. This is a much more challenging task than what we have achieved, and will require the development of more sophisticated models and data collection techniques.

Deployment performance would benefit from a lightweight model, and the model's proportion of inference time spent in convolution and LSTM suggests future work of static (for convolution) and dynamic (for LSTM) quantization. Optimized deployment to wearables would allow for timely feedback to be provided to users, helping them to optimize their training sessions and stop their sets with at most two reps in reserve.

\section{Conclusion}
This paper presented a novel model architecture for predicting muscular failure during preacher curls using motion data. This architecture is composed of an exercise repetition segmenter and near-failure classifier that achieved accuracies of 92.8\% and 86.6\%, respectively, despite a small dataset. The model was converted to CoreML for real-time use on Apple's iPhone 16 and Watch Series 10, demonstrating an average inference latency of 23.5 milliseconds, which is more than adequate for a user to take action based on the model's predictions.  The success of this model opens the door for future work on a regression model that predicts exact repetitions in reserve for multiple exercises, which could better assist users in applying different exercise strategies or a safer approach to muscular rehabilitation.

\section*{Acknowledgment}
This work is supported by the National Science Foundation (NSF) under grant number 2340249. The authors wish to acknowledge the reviewers and the study participants.

\bibliographystyle{IEEEtran}
\bibliography{references}

\end{document}